\documentclass[10pt,twocolumn,letterpaper]{article}

\usepackage{cvpr}
\usepackage{times}
\usepackage{epsfig}
\usepackage{graphicx}
\usepackage{amsmath}
\usepackage{amssymb}

\usepackage[breaklinks=true,bookmarks=false]{hyperref}

\cvprfinalcopy 

\def\cvprPaperID{****} 

\begin{document}

\title{Video-based Person Re-identification via 3D Convolutional Networks\\ and Non-local Attention}

\author{Xingyu Liao\footnotemark[1] \ \textsuperscript{1} \ , Lingxiao He\footnotemark[1] \ \textsuperscript{2} \ , Zhouwang Yang\textsuperscript{1} \ Chi Zhang\textsuperscript{3} \\
\textsuperscript{1}University of Science and Technology of China, Hefei, P.R.China \\ \textsuperscript{2}University of Chinese Academy of Sciences, Beijing, P.R.China\\ \textsuperscript{3} Megvii Inc.(Face++) \\
{\tt\small randall@mail.ustc.edu.cn, lingxiao.he@nlpr.ia.ac.cn, yangzw@ustc.edu.cn, zhangchi@megvii.com}
}
\renewcommand{\thefootnote}{\fnsymbol{footnote}}

\maketitle
\footnotetext[1]{Authors contributed equally.}

\begin{abstract}
   Video-based person re-identification (ReID) is a challenging problem,
   where some video tracks of people across non-overlapping cameras are available for matching.  Feature aggregation from a video track is a key step for video-based person ReID. Many existing methods tackle this problem by average/maximum temporal pooling or RNNs with attention. However, these methods cannot deal with temporal dependency and spatial misalignment problems at the same time. We are inspired by video action recognition that involves the identification of different actions from video tracks. Firstly, we use 3D convolutions on video volume, instead of using 2D convolutions across frames, to extract spatial and temporal features simultaneously. Secondly, we use a non-local block to tackle the misalignment problem and capture spatial-temporal long-range dependencies. As a result, the network can learn useful spatial-temporal information as a weighted sum of the features in all space and temporal positions in the input feature map. Experimental results on three datasets show that our framework outperforms state-of-the-art approaches by a large margin on multiple metrics.
\end{abstract}

\section{Introduction}

Person re-identification (ReID) aims to match people in the different places (time) using another non-overlapping camera, which has become increasingly popular in recent years due to the wide range of applications, such as public security, criminal investigation, and surveillance. Most deep learning approaches have been shown to be more effective than traditional methods \cite{Ahmed,Chen,Nips_huang,PR_ding}. But there still remains many challenging problems because of human pose, lighting, background, occluded body region and camera viewpoints.

Video-based person ReID approaches consist of feature representation and feature aggregation. And feature aggregation attracts more attention in recent works. Although most of methods \cite{jiyanggao} (see Fig. 1(A)) propose to use average or maximum temporal pooling to aggregate features, they do not take full advantage of the temporal dependency information. To this end, RNN based methods \cite{McLaughlin_2016_CVPR} (see Fig. 1(B)) are proposed to aggregate the temporal information among video frames. However, the most discriminative frames cannot be learned by RNN based methods while treating all frames equally. Moreover, temporal attention methods \cite{Liu_2017_CVPR} as shown in Fig. 1(C) are proposed to extract the discriminative frames. In conclusion, these methods mentioned above cannot tackle temporal dependency, attention and spatial misalignment simultaneously. Although there are a few methods \cite{Xu_2017_ICCV} using the jointly attentive spatial-temporal scheme, it is hard to optimize the networks under severe occlusion.

\begin{figure}[t]
	\centering
	\includegraphics[width=8.5cm]{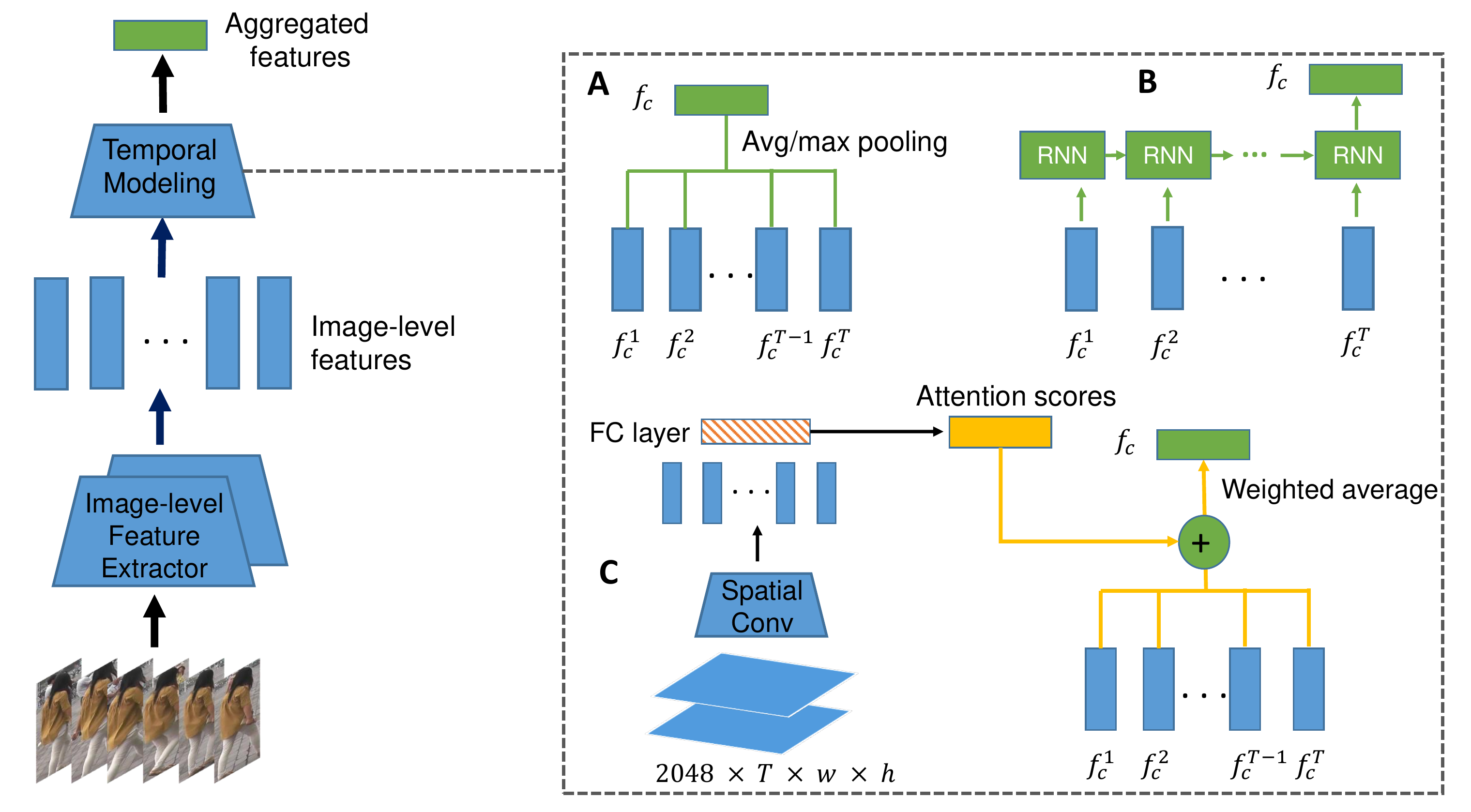}
	\caption{Three temporal modeling methods (A: temporal pooling, B: RNN, C: temporal attention) based on an image-level feature extractor (typically a 2D CNN). For temporal pooling, average or maximum pooling is used. For RNN, hidden state is used as the aggregated feature. For attention, spatial conv + FC is shown.}
	\label{fig:example}
\end{figure}

In this paper, we propose a method to aggregate temporal-dependency features and tackle spatial misalignment problems using attention simultaneously as illustrated in Fig. 2. Inspired by the recent success of 3D convolutional neural networks on video action recognition \cite{3dconv,i3d}, we directly use it to extract spatial-temporal features in a sequence of video frames. It can integrate feature extraction and temporal modeling into one step. In order to capture long-range dependency, we embed the non-local block \cite{Wang_2018_CVPR,Dai_2017_ICCV} into the model to obtain an aggregate spatial-temporal representation. We summarize the contributions of this work in three-folds.

\begin{figure}[t]
	\centering
	\includegraphics[width=8cm]{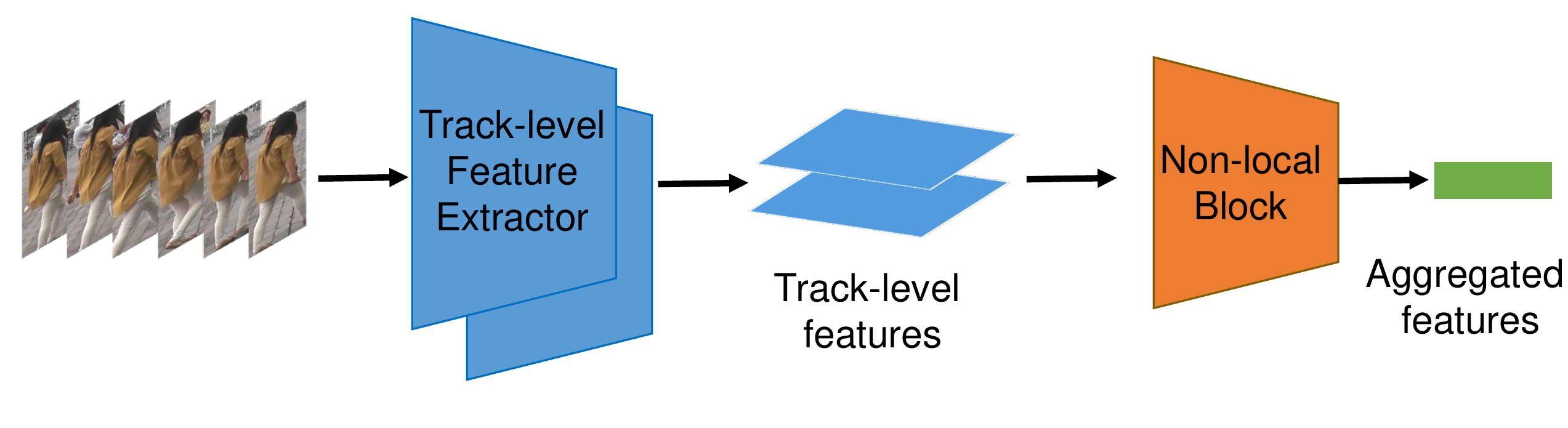}
	\caption{The overall architecture of the proposed method. 3D convolutions are used for track-level feature extractor. Non-local blocks are embedded into aggregate spatial-temporal features.}
	\label{fig:example}
\end{figure}
\begin{enumerate}
	\item
	We first propose to use 3D convolutional neural network to extract the aggregate representation of spatial-temporal features, which is capable of discovering pixel-level information and relevance among video tracks.
	\item	
	Non-local block, as a spatial-temporal attention strategy, explicitly solves the misalignment problem of deformed images. Simultaneously, the aggregative feature can be learned from video tracks by the temporal attentive scheme.
	\item
	Spatial attention and temporal attention are incorporated into an end-to-end 3D convolution model, which achieves significant performance compared to the existing state-of-the-art approaches on three challenging video-based ReID datasets.
\end{enumerate}

The rest of this paper is organized as follows. In Section 2, we discuss some related works. Section 3 introduces the details of the proposed approach. Experimental results on three public datasets will be given in Section 4. At last, we conclude this paper in Section 5.

\section{Related Work}

In this section, we first review some related works in person ReID, especially those video-based methods. Then we will discuss some related works about 3D convolution neural networks and non-local methods.

\subsection{Person Re-ID}

\noindent\textbf{Image-based person ReID} mainly focuses on feature fusion and alignment with some external information such as mask, pose, and skeleton, etc. Zhao \textit{et al.} \cite{Zhao_2017_CVPR} proposed a novel Spindle Net based on human body region guided multi-stage feature decomposition and tree-structured competitive feature fusion. Song \textit{et al.} \cite{Song_2018_CVPR} introduced the binary segmentation masks to construct synthetic RGB-Mask pairs as inputs, as well as a mask-guided contrastive attention model (MGCAM) to learn features separately from body and background regions. Suh \textit{et al.} \cite{sun_part_aligned} proposed a two-stream network that consists of appearance map extraction stream and body part map extraction stream, additionally a part-aligned feature map is obtained by a bilinear mapping of the corresponding local appearance and body part descriptors. These models all actually solve the person misalignment problem.

\noindent\textbf{Video-based person ReID} is an extension of image-based methods. Instead of pairs of images, the learning algorithm is given pairs of video sequences. The most important part is how to fuse temporal features from video tracks. Wang \textit{et al.} \cite{wang} aimed at selecting discriminative spatial-temporal feature representations. They firstly choosed the frames with the maximum or minimum flow energy, which is computed by optical flow fields. In order to take full use of temporal information, McLaughlin \textit{et al.} \cite{McLaughlin_2016_CVPR} built a CNN to extract features of each frame and then used RNN to integrate the temporal information between frames, the average of RNN cell outputs are adapted to summarize the output feature. Similar to \cite{McLaughlin_2016_CVPR}, Yan \textit{et al.} \cite{Yan_rnn} also used RNNs to encode video tracks into sequence features, the final hidden state is used as video representation. RNN based methods treat all frames equally, which cannot focus on more discriminative frames.Liu \textit{et al.} \cite{Liu_2017_CVPR} designed a Quality Aware Network (QAN), which is essentially an attention weighted average, to aggregate temporal features; the attention scores are generated from frame-level feature maps. In 2016, Zheng \textit{et al.} \cite{zheng2016mars} built a new dataset MARS for video-based person ReID, which becomes the standard benchmark for this task.

\subsection{3D ConvNets}
3D CNNs are well-suited for spatial-temporal feature learning. Ji \textit{et al.} \cite{3dconv} first proposed a 3D CNN model for action recognition. Tran \textit{et al.} \cite{Tran_2015_ICCV} proposed a C3D network to be applied into various video analysis tasks. Despite 3D CNNs' ability to capture the appearance and motion information encoded in multiple adjacent frames effectively, it is difficult to be trained with more parameters. More recently, Carreira \textit{et al.} \cite{i3d} proposed the Inflated 3D (I3D) architecture which initializes the model weights by inflating the pre-trained weights from ImageNet \cite{ILSVRC15} over temporal dimension which significantly improves the performance of 3D CNNs and it is the current state-of-the-art on the Kinetics dataset \cite{kinetics}.
\begin{figure*}[t]
	\centering
	\includegraphics[width=17cm]{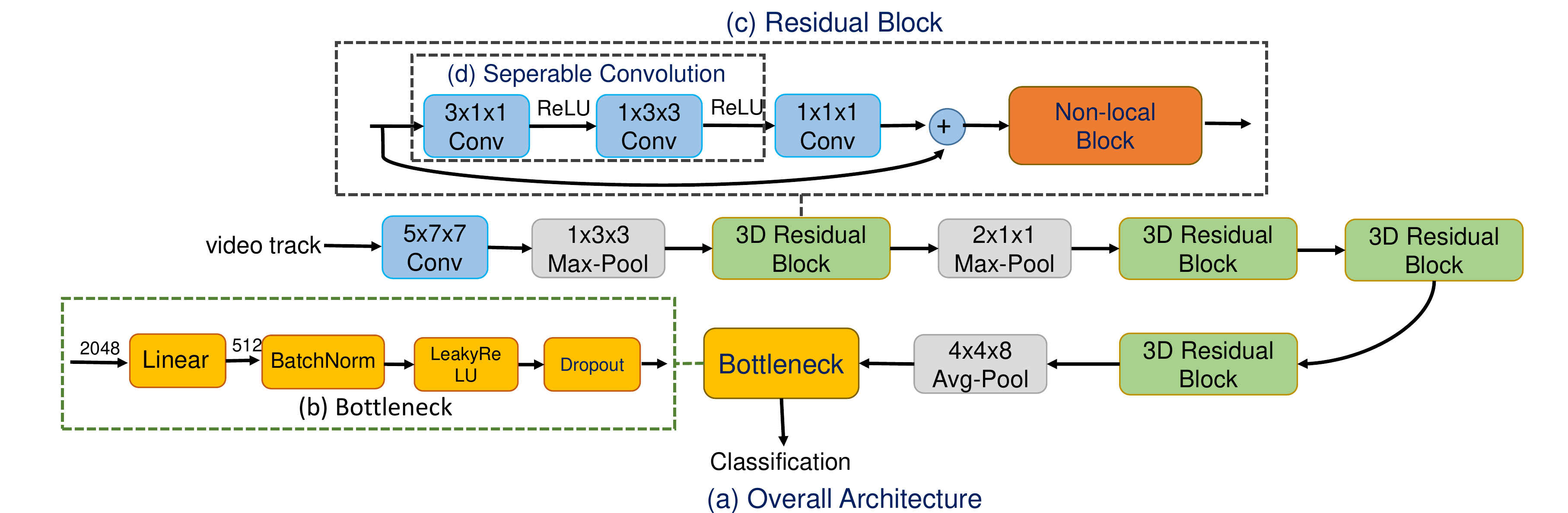}
	\caption{Illustration of networks we propose in this paper; (a) illustrates the overall architecture that is consist of 3D convolutions, 3D pooling, 3D residual blocks, bottleneck and non-local blocks; (b) shows bottelneck; (c) illustrates residual blocks; Seperable convolutions are shown in (d).}
	\label{fig:example}
\end{figure*}

\subsection{Self-attention and Non-local}

Non-local  technique \cite{non-local-image} is a classical digital image denoising algorithm that computes a weighted average of all pixels in an image. As attention models grow in popularity, Vaswani \textit{et al.} \cite{NIPS2017_attention} proposed a self-attention method for machine translation that computes the response at a position in a sequence (\textit{e.g.,} a sentence) by attending to all positions and taking their weighted average in an embedding space. Moreover, Wang \textit{et al.} \cite{Wang_2018_CVPR} proposed a non-local architecture to bridge self-attention in machine translation to the more general class of non-local filtering operations. Inspired by these works, We embed non-local blocks into I3D model to capture long-range dependencies on space and time for video-based ReID. Our method demonstrates better performance by aggregating the discriminative spatial-temporal features.

\section{The Proposed Approach}

In this section, we introduce the overall system pipeline and detailed configurations of the spatial-temporal modeling methods. The whole system could be divided into two important parts: extracting spatial-temporal features from video tracks through 3D ResNet, and integrating spatial-temporal features by the non-local blocks.

A video tracet is first divided into consecutive non-overlap tracks $ \{ c_k \} $, and each track contains $N$ frames. Supposing each track is represented as

\begin{equation}
	c_k = \{x_t | x_t \in \mathbb{R}^{H \times W} \}_{t=1}^N ,
\end{equation}
where $N$ is the length of $c_k$, and $H$, $W$ are the height, width of the images respectively. As shown in Fig. 3(a), the proposed method directly accepts a whole video track as the inputs and outputs a $d$-dimensional feature vector $f_{c_k}$. At the same time, non-local blocks are embedded into 3D residual block (Fig. 3(c)) to integrate spatial and temporal features, which can effectively learn the pixel-level relevance of each frame and learn hierarchical feature representation.

Finally, average pooling followed by a bottleneck block (Fig. 3(b)) to speed up training and improve performance. A fully-connected layer is added on top to learn the identity features. A Softmax cross-entropy with label smoothing, proposed by Szegedy \textit{et al.} \cite{Szegedy_2016_CVPR}, is built on top of the fully connected layer to supervise the training of the whole network in an end-to-end fashion. At the same time, Batch Hard triplet loss \cite{triplet} is employed in the metric learning step. During the testing, the final similarity between $ c_i $ and $ c_j $ can be measured by L2 distance or any other distance function.

In the next parts, we will explain each important component in more detail.

\subsection{Temporally Separable Inflated 3D Convolution}

In 2D CNNs, convolutions are applied on the 2D feature maps to compute features from the spatial dimensions only. When applied to the video-based problem, it is desirable to capture the temporal information encoded in multiple contiguous frames. The 3D convolutions are achieved by convolving 3D kernel on the cube formed by stacking multiple consecutive frames together. In other words, 3D convolutions can directly extract a whole representation for a video track, while 2D convolutions first extract a sequence of image-level features and then features are aggregated into a single vector feature. Formally, the value at position $ (x, y, z) $ on the $j$-th feature map in the $i$th layer $V_{ij}^{xyz}$ is given by

\begin{equation}
	V_{ij}^{xyz} = b_{ij} + \sum_m \sum_{p=0}^{P_i - 1} \sum_{q=0}^{Q_i - 1} \sum_{r=0}^{R_i - 1} W_{ijm}^{pqr} V_{(i - 1)m}^{(x + p) (y+q)(z+r)},
\end{equation}
where $P_i$ and $Q_i$ are the height and width of the kernel, $R_i$ is the size of the 3D kernel along with the temporal dimension, $W_{ijm}^{pqr}$ is the $(p, q, r) $th value of the kernel connected to the $m$-th feature map in the previous layer $V_{i - 1}$, and $b_{ij}$ is the bias.

We adopt 3D ResNet-50 \cite{resnet} that uses 3D convolution kernels with ResNet architecture to extract spatial-temporal features. However, C3D-like 3D ConvNet \cite{DuTran} is hard to optimize because of a large number of parameters. In order to address this problem, we inflate all the 2D ResNet-50 convolution filters with an additional temporal dimension. For example, a 2D $k \times k$ kernel can be inflated as a 3D $t \times k \times k$ kernel that spans $t$ frames. We initialize all 3D kernels with 2D kernels (pre-trained on ImageNet): each of the $t$ planes in the $t \times k \times k$ kernel is initialized by the pre-trained $k \times k$ weights, rescaled by $1 / t$. According to Xie \textit{et al.} \cite{xie2017rethinking} experiments, temporally separable convolution is a simple way to boost performance on variety of video understanding tasks. We replace 3D convolution with two consecutive convolution layers: one 1D convolution layer purely on the temporal axis, followed by a 2D convolution layer to learn spatial features in Residual Block as shown in Fig. 3(d). Meanwhile, we pre-train the 3D ResNet-50 on Kinetics \cite{kinetics} to enhance the generalization performance of the model. We replace the final classification layer with person identity outputs. The model takes $T$ consecutive frames (\textit{i.e.} a video track) as the input, and the layer outputs before final classification layer is used as the video track identity representation.

\subsection{Non-local Attention Block}

A non-local attention block is used to capture long-range dependency in space and time dealing with occlusion and misalignment. We first give a general definition of non-local operations and then provide the 3D non-local block instantiations embedded into the I3D model.

Following the non-local methods \cite{non-local-image} and \cite{Wang_2018_CVPR}, the generic non-local operation in deep neural networks can be given by
\begin{equation}
\begin{array}{l}
\displaystyle y_i = \frac{1}{\mathcal{C}(x)} \sum_{\forall j} f(x_i, x_j) g(x_j).
\end{array}
\label{eq:6}
\end{equation}
Here $x_i$ can be the position in input signal (image, sequence, video; often their features) and $y_i$ is the position in output signal of the same size as $x$, whose response is to be computed by all possible input positions $x_j$.  A pairwise function $f$ computes a scalar between $i$ and all $j$, which represents attention scores between position $i$ in output feature and all position $j$ in the input signal. The unary function $g$ computes a representation in an embedded space of the input signal at the position $j$. At last, the response is normalized by a factor $\mathcal{C}(x)$ as shown in Fig. 4.

\begin{figure}[t]
	\centering
	\includegraphics[width=8cm]{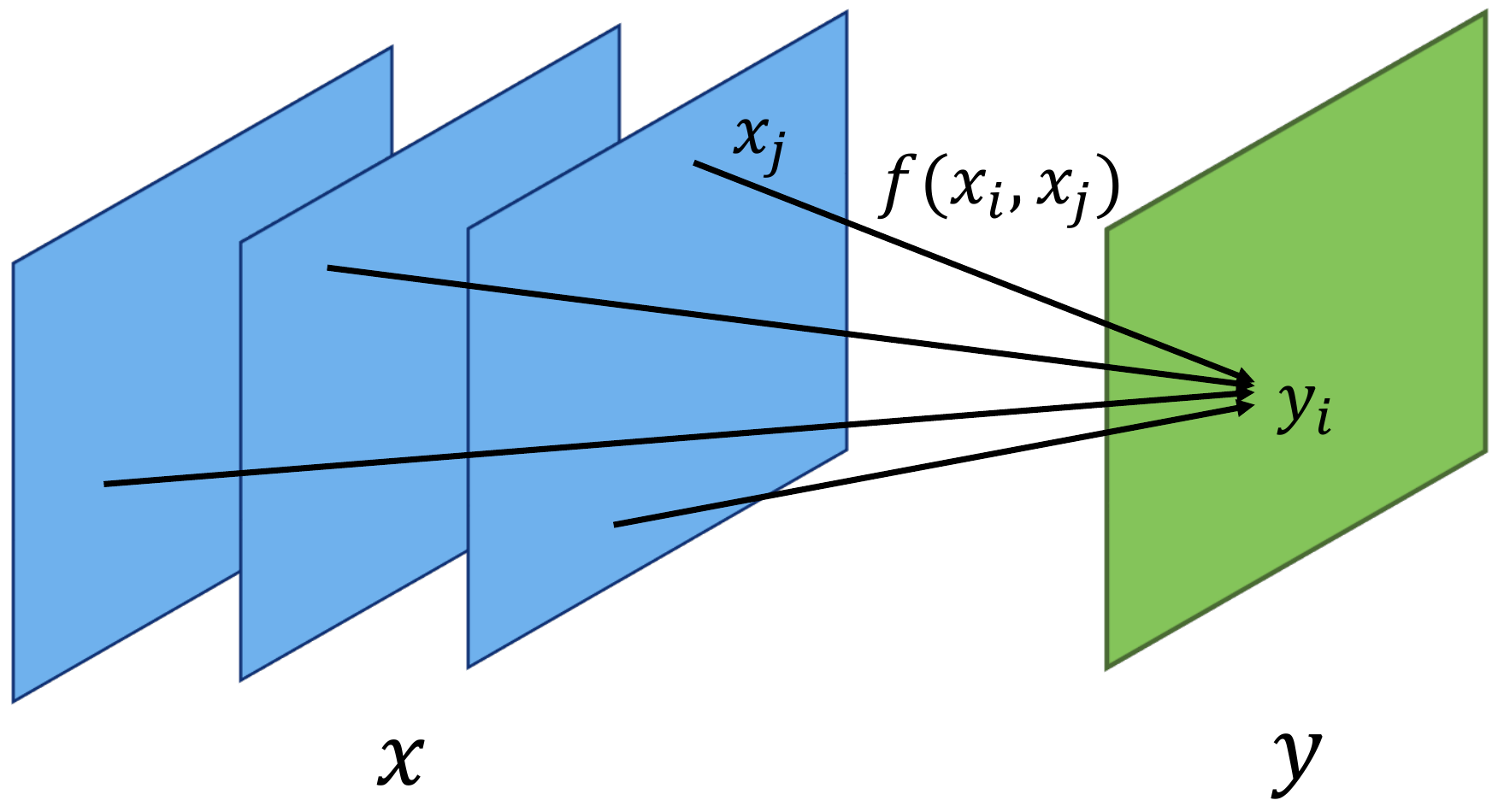}
	\caption{An illustration of generic non-local methods. $x$ is input signal (image, sequence, video or their features). $y$ is output signal.}
	\label{fig:example}
\end{figure}

Because of the fact that all positions $( \forall j)$ are considered in the operation in Eq.(2), this is so-called non-local. Compared with this, a standard 1D convolutional operation sums up the weighted input in a \textit{local} neighborhood (\textit{e.g.}, $i-1 \leq j \leq i+1$ with kernel size 3, and recurrent operation at time $i$ is often based only on the current and the latest time step (\textit{e.g.}, $j=i$ or $i - 1$).

There are several versions of $f$ and $g$, such as gaussian, embedded gaussian, dot product, etc. According to experiments in \cite{Wang_2018_CVPR}, the non-local operation is not sensitive to these choices. We just choose embedded gaussian as $f$ function that is given by

\begin{equation}
	f(x_i, x_j) = e^{\theta (x_i)^T \phi (x_j)}
\end{equation}
Here $x_i$ and $x_j$ are given in Eq. (3), $\theta (x_i) = W_{\theta} x_i$ and $\phi (x_j) = W_{\phi} x_j$ are two embeddings.  We can set $\mathcal{C} (x)$ as a softmax operation, so we have a self-attention form that is given by

\begin{equation}
	y =  \sum_{\forall j} \frac{e^{\theta (x_i)^T \phi (x_j)}}{\sum_{\forall i} e^{\theta (x_i)^T \phi (x_j)}} g(x_j)
\end{equation}

A non-local operation is very flexible, which can be easily incorporated into any existing architecture. The non-local operation can be wrapped into a non-local block that can be embedded into the earlier or later part of the deep neural network. We define a non-local block as:
\begin{equation}
	z_i = W_z y_i + x_i
\end{equation}
where $y_i$ is given in Eq.(3) and "$+x_i$" means a residual connection \cite{resnet}. We can plug a new non-local block into any pre-trained model, without breaking its initial behavior (\textit{e.g.}, if $W_z$ is initialized as zero) which can build a richer hierarchy architecture combining both global and local information.

In ResNet3D-50, we use a 3D spacetime non-local block illustrated in Fig. 5. The pairwise computation in Eq.(4) can be simply done by matrix multiplication. We will talk about detailed implementation of non-local blocks in next part.

\begin{figure}
	\centering
	\includegraphics[width=8cm]{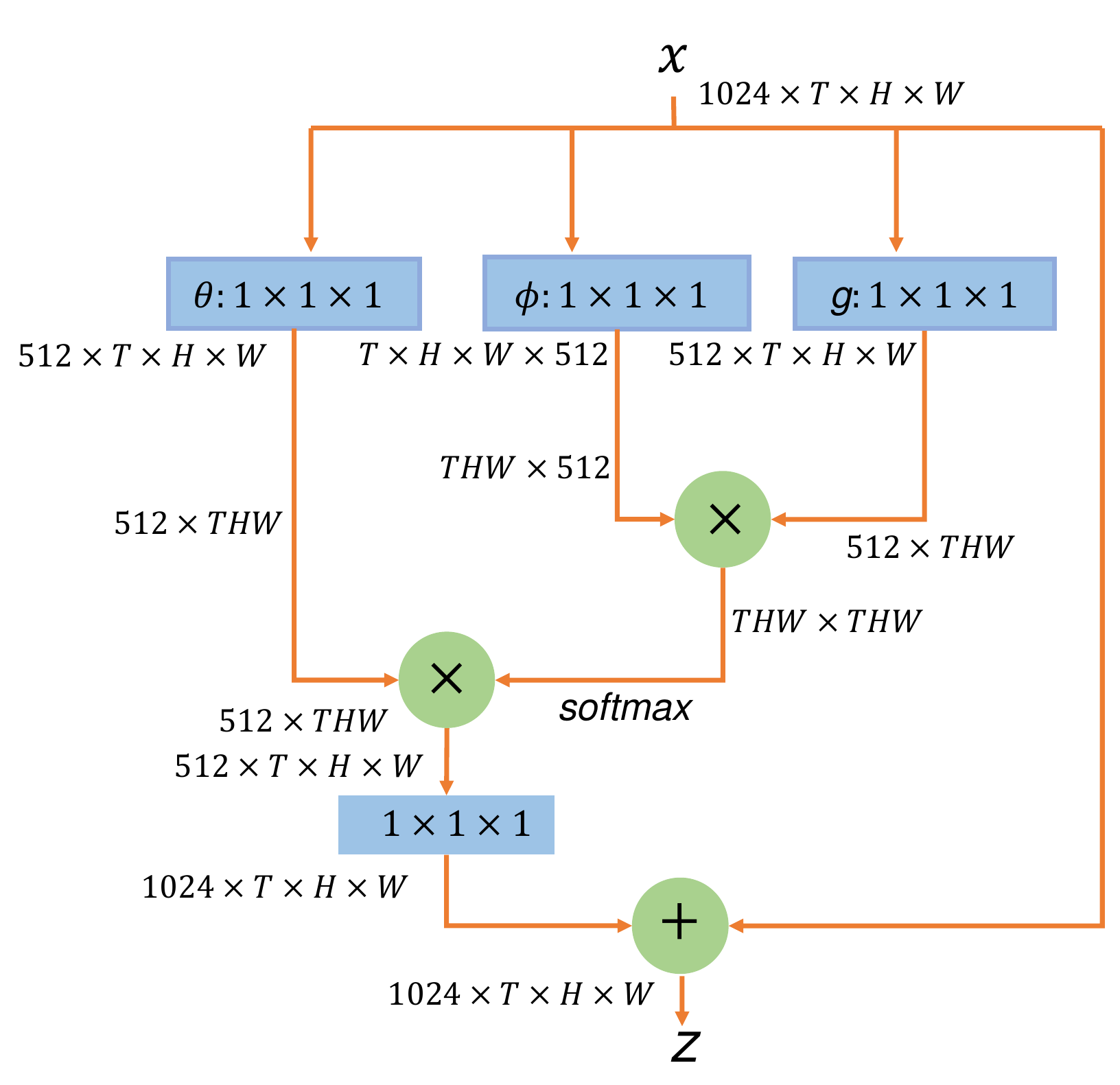}
	\caption{The 3D spacetime non-local block. The feature maps are shown as the shape of their tensors, \textit{e.g.}, $1024 \times T \times H \times W$ for 1024 channels (it can be different depending on networks). "$\otimes$" denotes matrix multiplication, and "$\oplus$" denotes element-wise sum. The softmax operation is performed on each row. The blue boxes denote $1 \times 1 \times 1$ convolutions. We show the Embedded Gaussian version, with a bottleneck of 512 channels.}
	\label{fig:example}
\end{figure}

\subsection{Loss Functions}

We use triplet loss function with hard mining \cite{triplet} and a Softmax cross-entropy loss function with label smoothing regularization \cite{Szegedy_2016_CVPR}.

The triplet loss function we use was originally proposed in \cite{triplet}, and named as Batch Hard triplet loss function. To form a batch, we randomly sample $P$ identities and randomly sample $K$ tracks for each identity (each track contains $T$ frames); totally there are $P \times K$ clips in a batch. For each sample $a$ in the batch, the hardest positive and the hardest negative samples within the batch are selected when forming the triplets for computing the loss $L_{triplet}$.
\begin{equation}
	\begin{aligned}
		L_{triplet} = \overbrace{\sum_{i=1}^P \sum_{a=1}^K}^{all\ anchors} [m + \overbrace{\max \limits_{p=1 \cdots K} D(f_{a}^i, f_{p}^i)}^{hardest\ positive} \\ - \underbrace{\min \limits_{\substack{j = 1 \cdots P \\ n = 1\cdots K \\ j \neq i}} D(f_{a}^i, f_{n}^j) }_{hardest \ negative}]_{+}
	\end{aligned}
\end{equation}

The original Softmax cross-entropy loss function is given by:

\begin{equation}
	L_{softmax} = -\frac{1}{P \times K} \sum_{i=1}^P \sum_{a=1}^K p_{i,a} \log q_{i, a}
\end{equation}
where $p_{i,a}$ is the ground truth identity and $q_{i,a}$ is prediction of sample $\{ i, a\}$. The \textit{label-smoothing regularization} is proposed to regularize the model and make it more adaptable with:
\begin{equation}
	L^{'}_{softmax} = -\frac{1}{P \times K} \sum_{i=1}^P \sum_{a=1}^K p_{i,a} \log ( (1 - \epsilon) q_{i, a} + \frac{\epsilon}{N})
\end{equation}

where $N$ is the number of classes. This can be considered as a mixture of the original ground-truth distribution $ q_{i, a}$ and the uniform distribution $u(x) = \frac{1}{N}$.

The total loss L is the combination of these two losses.

\begin{equation}
	L = L^{'}_{softmax} + L_{triplet}
\end{equation}

\section{Experiments}

We evaluate our proposed method on three public video datasets, including iLIDS-VID \cite{wang}, PRID-2011 \cite{prid} and MARS \cite{zheng2016mars}. We compare our method with the state-of-the-art methods, and the experimental results demonstrate that our proposed method can enhance the performance of both feature learning and metric learning and outperforms previous methods.

\subsection{Datasets}

The basic information of three dataset is listed in Table 1 and some samples are displayed in Figure [3].

\renewcommand{\arraystretch}{1.1}
\begin{table}
\centering
\footnotesize
\caption{The basic information of three datasets to be used in our experiments.}
	\begin{tabular}{llll}
\hline
		Datasets & iLIDS-VID & PRID2011 & MARS\\
			
			\hline
			$\#$identities  & 300   & 200 & 1,261 \\
			$\#$track-lets & 600  & 400 & 21K  \\
			$\#$boxes & 44K  & 40K & 1M \\
			$\#$distractors & 0  & 0 & 3K \\
			$\#$cameras & 2  & 2 & 6 \\
			$\#$resolution & $64 \times 128$  & $64 \times 128$ & $128 \times 256$ \\
			$\#$detection & hand  & hand & algorithm \\
			$\#$evaluation & CMC & CMC & CMC $\&$ mAP \\
			\hline
	\end{tabular}
\end{table}

\noindent\textbf{iLIDS-VID} dataset consists of 600 video sequences of 300 persons. Each image sequence has a variable length ranging from 23 to 192 frames, with averaged number of 73. This dataset is challenging due to clothing similarities among people and random occlusions.

\noindent\textbf{PRID-2011} dataset contains 385 persons in camera A and 749 in camera B. 200 identities appear in both cameras, constituting of 400 image sequences. The length of each image sequence varies from 5 to 675. Following \cite{zheng2016mars}, sequences with more 21 frames are selected, leading to 178 identities. 

\noindent\textbf{MARS} dataset is a newly released dataset consisting of 1,261 pedestrians captured by at least 2 cameras. The bounding boxes are generated by classic detection and tracking algorithms (DPM detector) \cite{dpm}, yielding 20,715 person sequences. Among them, 3,248 sequences are of quite poor quality due to the failure of detection or tracking, significantly increasing the difficulty of person ReID.

\subsection{Implementation Details and Evaluation Metrics}

\textbf{Training.} We use ResNet3D-50 \cite{resnet3d} as our backbone network. According to the experiments in \cite{Wang_2018_CVPR}, five non-local blocks are inserted to right before the last residual block of a stage. Three blocks are inserted into $res_4$ and two blocks are inserted into $res_3$, to every other residual block. Our models are pre-trained on Kinetics \cite{kinetics}; we also compare the models with different pre-trained weights, and the details are described in the next section.

Our implementation is based on publicly available code of PyTorch \cite{pytorch}. All person ReID models in this paper are trained and tested on Linux with GTX TITAN X GPU. In training term, eight-frame input tracks are randomly cropped out from 64 consecutive frames every eight frames. The spatial size is $256 \times 128$ pixels, randomly cropped from a scaled videos whose size is randomly enlarged by 1/8. The model is trained on an eight-GPU machine and each GPU have 16 tracks in a mini-batch (so in total with a mini-batch size of 128 tracks). In order to train hard mining triplet loss, 32 identities with 4 tracks each person are taken in a mini-batch and iterate all identities as an epoch. Bottleneck consists of fully connected layer, batch norm, leaky ReLU with $\alpha=0.1$ and dropout with $0.5$ drop ratio. The model is trained for 300 epochs in total, starting with a learning rate of 0.0003 and reducing it by exponential decay with decay rate 0.001 at 150 epochs. Adaptive Moment Estimation (Adam) \cite{adam} is adopted with a weight decay of 0.0005 when training.

The method in \cite{kaiming} is adopted to initialize the weight layers introduced in the non-local blocks. A BatchNorm layer is added right after the last $1 \times 1 \times 1$ layer that represents $W_z$; we do not add BatchNorm to other layers in a non-local block. The scale parameter of this BatchNorm layer is initialized as zeros, following \cite{large_minibatch}. This ensures that the initialize state of the entire non-local block is an identity mapping, so it can be inserted into any pre-trained networks while maintaining its initial behavior.
\renewcommand{\arraystretch}{1.1}
\begin{table}
\centering
\footnotesize
		\caption{Component analysis of the proposed method: rank-1, rank-5, rank-10 accuracies and mAP are reported for MARS dataset. \textbf{ResNet3D-50} is the ResNet3D-50 pre-trained on Kinectis, \textbf{ResNet3D-50 NL} is added with non-local blocks.}
\vspace{0.5em}
		\label{table:headings}
		\begin{tabular}{lllll}
			Methods & CMC-1 & CMC-5 & CMC-10 & mAP\\
			\hline
			\textbf{Baseline}  & 77.9   & 90.0  & 92.5  & 69.0   \\
			\textbf{ResNet3D-50} & 80.0   & 92.2  & 94.5  & 72.6   \\
			\textbf{ResNet3D-50 NL} & 84.3  & 94.6  & 96.2  & 77.0   \\
			\hline
		\end{tabular}
\end{table}

\textbf{Testing.} We follow the standard experimental protocols for testing on the datasets. For iLIDS-VID, the 600 video sequences of 300 persons are randomly split into $50 \%$ of persons for testing. For PRID2011, only 400 video sequences of the first 200 persons, who appear in both cameras are used  according to experiment setup in previous methods \cite{McLaughlin_2016_CVPR} For MARS, the predefined 8,298 sequences of 625 persons are used for training, while the 12,180 sequences of 636 persons are used for testing, including the 3,248 low quality sequences in the gallery set.

We employ Cumulated Matching Characteristics (CMC) curve and mean average precision (mAP) to evaluate the performance for all the datasets. For ease of comparison, we only report the cumulated re-identification accuracy at selected ranks.

\begin{figure}
	\centering
	\includegraphics[width=1\linewidth]{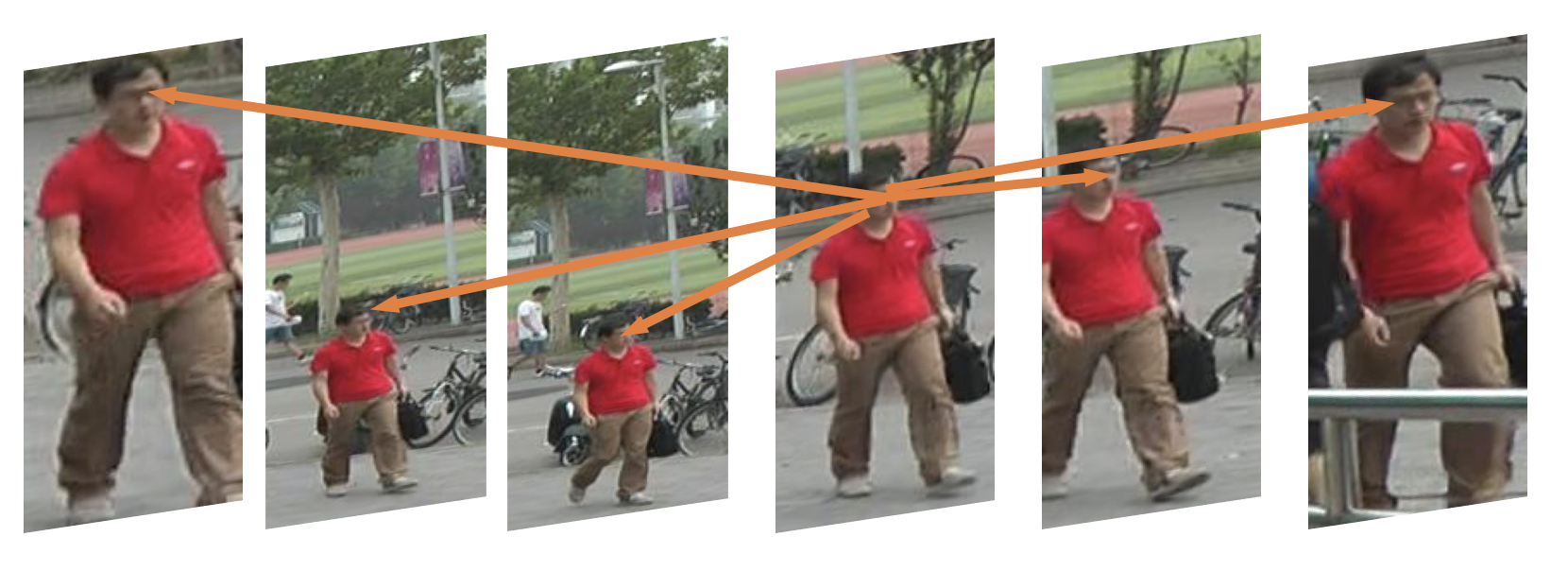}
	\caption{Example of the behavior of a non-local block to tackle misalignment problems. The starting point of arrows represents one $x_i$, and the ending points represent $x_j$. This visualization shows how the model finds related part on different frames.}
	\label{fig:example}
\end{figure}

\subsection{Component Analysis of the Proposed Model}
In this part, we report the performance of different components in our models.
\subsubsection{3D CNN and Non-local.} Baseline method, ResNet3D-50 and ResNet3D-50 with non-local blocks on the MARS dataset are shown in Table 2. \textbf{Baseline} corresponds to ResNet-50 trained with softmax cross-entropy loss and triplet with hard mining on image-based person ReID. The representation of an image sequence is obtained by using the average temporal pooling. \textbf{ResNet3D-50} corresponds to ResNet3D-50 pre-trained on Kinetics discussed above. \textbf{ResNet3D-50 NL} corresponds to ResNet3D-50 with non-local blocks pre-trained on Kinetics. The gap between our results and baseline method is significant, and it is noted that: (1) ResNet3D increases from $77.9\%$ to $80.0\%$ under single query, which fully suggests ResNet3D-50 effectively aggregate the spatial-temporal features; (2) ResNet3D with non-local increase from $80.0 \%$ to $84.3 \%$ compared with ResNet3D, which indicates that non-local blocks have the great performance on integrating spatial-temporal features and tackling misalignment problem. The results are shown in Fig. 6.

\renewcommand{\arraystretch}{1.1}
\begin{table}
\centering
\footnotesize
		\caption{
			Effect of different initialization methods: rank-1, rank-5, rank-10 accuracies and mAP are reported for MARS dataset. \textbf{ImageNet} corresponds to model pre-trained on ImageNet, \textbf{Kinetics} corresponds to model pre-trained on Kinetics and \textbf{ReID} corresponds to model pre-trained on ReID datasets.
		}
		\label{table:headings}
		\begin{tabular}{lllll}
			\hline\noalign{\smallskip}
			Init Methods & CMC-1 & CMC-5 & CMC-10 & mAP\\
			\noalign{\smallskip}
			\hline
			\noalign{\smallskip}
			\textbf{ImageNet}  & 78.4   &91.5 & 93.9  & 69.8   \\
			\textbf{ReID} &  79.9  &  92.6 & 94.5 & 71.3 \\
			\textbf{Kinetics} & 84.3    & 94.6  & 96.2  & 77.0   \\
			\hline
		\end{tabular}
\end{table}

\renewcommand{\arraystretch}{1.1}
\begin{table}
\centering
\footnotesize
		\caption{
			Comparisons of our proposed approach to the state-of-the-art on PRID2011, iLIDS-VID and MARS datasets. The rank1 accuracies are reported and for MARS we provide mAP in brackets. The best and second best results are marked by {\color{red}{red}} and {\color{blue}{blue}} colors, respectively.
		}
		\label{table:headings}

		\begin{tabular}{llll}
			Methods   \qquad\qquad  & PRID2011 & iLIDS-VID & MARS \\
			\hline
			STA \cite{Liu_ICCV}  & 64.1  & 44.3 & - \\
			DVDL \cite{dvdl} & 40.6  & 25.9  & - \\
			TDL \cite{tdl} & 56.7  & 56.3  & - \\
			SI2DL \cite{si2dl} & 76.7  & 48.7  & - \\
			mvRMLLC+Alignment \cite{mvrm} & 66.8  & 69.1  & - \\
			AMOC+EpicFlow \cite{amoc} & 82.0  & 65.5  & - \\
			RNN \cite{McLaughlin_2016_CVPR} & 40.6  & 58.0  & - \\
			IDE \cite{ide} + XQDA \cite{xqda} & -  & -  & 65.3(47.3) \\
			end AMOC+epicFlow \cite{amoc} & 83.7  & 68.7  & 68.3(52.9) \\
			Mars \cite{zheng2016mars} & 77.3  & 53.0  & 68.3(49.3) \\
			SeeForest \cite{Zhou} & 79.4  & 55.2  & 70.6(50.7) \\
			QAN \cite{Liu_2017_CVPR} & 90.3  & 68.0  & - \\
			Spatialtemporal \cite{Li_2018_CVPR} & {\color{red}{93.2}}  & {\color{blue}{80.2}}  & \color{blue}{82.3(65.8)} \\ \hline
			\textbf{ours} & {\color{blue}91.2}  & {\color{red} 81.3}  & {\color{red}{84.3(77)}} \\
			\hline
		\end{tabular}
\end{table}

\subsubsection{Different Initialization Methods.} We also carry out experiments to investigate the effect of different initialization methods in Table 3. \textbf{ImageNet} and \textbf{ReID} corresponds to ResNet3D-50 with non-local block, whose weights are inflated from the 2D ResNet50 pre-trained on ImageNet or on CUHK03 \cite{cuhk03}, VIPeR \cite{viper} and DukeMTMC-reID \cite{duke} respectively. \textbf{Kinetics} corresponds to ResNet3D-50 with non-local blocks pre-trained on Kinetics. The results show that model pre-trained on Kinetics has the best performance than on other two datasets. 3D model is hard to train because of the large number of parameters and it needs more datasets to pre-train. Besides, the model pre-trained on Kinetics (a video action recognition dataset) is more suitable for video-based problem.

\subsection{Comparision with State-of-the-art Methods}

Table 4 reports the performance of our approach with other state-of-the-art techniques.

\subsubsection{Results on MARS.} MARS is the most challenging dataset (it contains distractor sequences and has a substantially larger gallery set) and our methodology achieves a significant increase in mAP and rank1 accuracy. Our method improves the state-of-the-art by $2.0 \%$ compared with the previous best reported results $82.3 \%$ from Li \textit{et al.} \cite{Li_2018_CVPR} (which use spatialtemporal attention). SeeForest \cite{Zhou} combines six spatial RNNs and temporal attention followed by a temporal RNN to encode the input video to achieve $70.6 \%$. In contrast, our network architecture is straightforward to train for the video-based problem. This result suggests our ResNet3D with non-local is very effective for video-based person ReID in challenging scenarios.

\subsubsection{Results on iLIDS-VID and PRID.} The results on the iLIDS-VID and PRID2011 are obtained by fine-tuning from the pre-trained model on the MARS. Li \textit{et al.} uses spatialtemporal attention to automatically discover a diverse set of distinctive body
parts which achieves $93.2 \%$ on PRID2011 and $80.2 \%$ on iLIDS-VID. Our proposed method achieves the comparable results compared with it by $91.2 \%$ on PRID2011 and $81.3 \%$ on iLIDS-VID. 3D model cannot achieve the significant improvement because of the size of datasets. These two datasets are small video person ReID datasets, which lead to overfitting on the training set.

\section{Conclusion}

In this paper, we have proposed an end-to-end 3D ConvNet with non-local architecture, which integrates a spatial-temporal attention to aggregate a discriminative representation from a video track. We carefully design experiments to demonstrate the effectiveness of each component of the proposed method. In order to discover pixel-level information and relevance between each frames, we employ a 3D ConvNets. This encourages the network to extract spatial-temporal features. Then we insert non-local blocks into model to explicitly solves the misalignment problem in space and time. The proposed method with ResNet3D and non-blocks outperforms the state-of-the-art methods in many metrics.

{\small
\bibliographystyle{ieee}
\bibliography{egbib}
}

\end{document}